\documentclass{article}

\usepackage[final]{nips_2017}
\usepackage[utf8]{inputenc} 
\usepackage[T1]{fontenc}    
\usepackage{textcomp}
\usepackage{hyperref}       
\usepackage{url}            
\usepackage{booktabs}       
\usepackage{amsfonts}       
\usepackage{nicefrac}       
\usepackage{microtype}      
\usepackage{amsmath}
\usepackage{wrapfig,booktabs}
\usepackage{color}
\usepackage{xcolor}
\usepackage{graphicx}
\usepackage{subfig}
\usepackage[font=small]{caption}
\usepackage{pdfpages}
\usepackage{natbib}
\setcitestyle{square,comma,numbers}

\usepackage{cprotect} \usepackage{multirow} \usepackage{collcell}
\usepackage{subfig} \usepackage{float} \usepackage{color}
\usepackage{multirow}
\usepackage[toc,page]{appendix}

\newcolumntype{L}[1]{>{\raggedright\let\newline\\\arraybackslash\hspace{0pt}}m{#1}}
\newcolumntype{C}[1]{>{\centering\let\newline\\\arraybackslash\hspace{0pt}}m{#1}}
\newcolumntype{R}[1]{>{\raggedleft\let\newline\\\arraybackslash\hspace{0pt}}m{#1}}

\title{Predicting Scene Parsing and Motion Dynamics \\in the Future}

\author{Xiaojie Jin$^1$, Huaxin Xiao$^2$, Xiaohui Shen$^3$, Jimei Yang$^3$, Zhe Lin$^3$ \\ \textbf{Yunpeng Chen}$^2$\textbf{,} \textbf{Zequn Jie}$^4$\textbf{,} \textbf{Jiashi Feng}$^2$\textbf{,} \textbf{Shuicheng Yan}$^{5,2}$\\
\small $^1$NUS Graduate School for Integrative Science and Engineering (NGS), NUS\\
\small$^2$Department of ECE, NUS \quad $^3$Adobe Research \quad $^4$Tencent AI Lab \quad $^5$Qihoo 360 AI
Institute}

\begin{document}
\maketitle
\begin{abstract}
The ability of predicting the future is important for intelligent systems, \textit{e.g.} autonomous vehicles and robots to plan early and make decisions accordingly. Future scene parsing and optical flow estimation are two key tasks that help agents better understand their environments as the former provides dense semantic information, \textit{i.e.} what objects will be present and where they will appear, while the latter provides dense motion information, \textit{i.e.} how the objects will move. In this paper, we propose a novel model to simultaneously predict scene parsing and optical flow in unobserved future video frames. To
our best knowledge, this is the first attempt in jointly predicting scene parsing and
motion dynamics. In particular, scene parsing enables structured motion prediction by decomposing optical flow into different groups while optical flow estimation brings reliable pixel-wise correspondence to scene parsing. By exploiting this mutually beneficial relationship, our model shows significantly better parsing and motion prediction results when compared to well-established baselines and individual prediction models on the large-scale Cityscapes dataset. In addition, we also demonstrate that our model can be used to predict the steering angle of the vehicles, which further verifies the ability of our model to learn latent representations of scene dynamics.
\end{abstract}

\section{Introduction} Future prediction is an important problem for artificial
intelligence. To enable intelligent systems like autonomous vehicles and
robots to react to their environments, it is necessary to endow them with the
ability of predicting what will happen in the near future and plan accordingly,
which still remains an open challenge for modern artificial vision systems.

In a practical visual navigation system, scene parsing and dense motion estimation are two essential components for understanding the scene
environment. The former provides pixel-wise prediction of semantic categories
(thus the system understands what and where the objects are)
and the latter describes dense motion trajectories (thus the system learns how
the objects move). The visual system becomes ``smarter'' by leveraging the
prediction of these two types of information, \textit{e.g.} predicting how the
car coming from the opposite direction moves to plan the path ahead of time and
predict/control the steering angle of the
vehicle.
Despite numerous models have been proposed on scene
parsing~\cite{chen2014semantic,farabet2013learning,fullyconvseg,roy2014scene,schwing2015fully,socher2011parsing,mpf}
and motion
estimation~\cite{baker2011database,fouhey2014predicting,mester2014motion}, most
of them focus on processing observed images, rather than predicting in
unobserved future scenes. Recently, a few works~\cite{s2s,jinme,chao2017forecasting} explore how to
anticipate the scene parsing or motion dynamics, but they all tackle these two
tasks separately and fail to utilize the benefits that one task brings to the
other.

\begin{figure}[t!]
\captionsetup[subfigure]{labelformat=empty}
	\centering
	\subfloat[]{
	    \includegraphics[width=\linewidth]{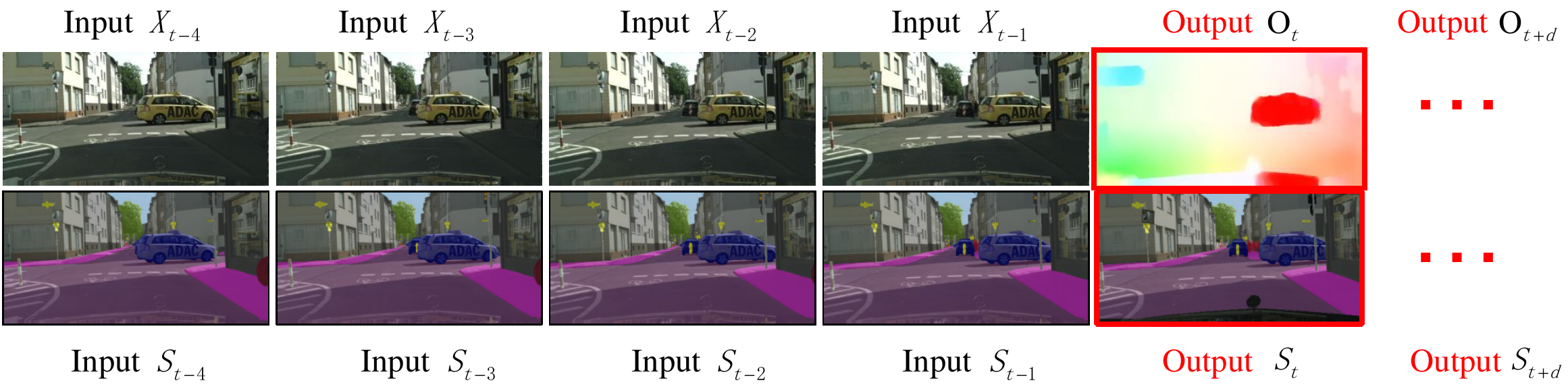}}
          \caption{Our task. The proposed model jointly predicts scene
            parsing and optical flow in the future. \textbf{Top}: Future flow
            (highlighted in red) anticipated using preceding
            frames. \textbf{Bottom}: Future scene parsing (highlighted in red)
            anticipated using preceding scene parsing results. We use the flow
            field color coding from~\cite{baker2011database}. }
    \label{fig:goal}
\end{figure}

In this paper, we try to close this research gap by presenting a novel model for
jointly predicting scene parsing and motion dynamics (in terms of the
dense optical flow) for future frames. More importantly, we leverage one task as the auxiliary of
the other in a mutually boosting way. See Figure~\ref{fig:goal} for an
illustration of our task. For the task of predictive scene parsing, we use the
discriminative and temporally consistent features learned in motion prediction to
produce parsing prediction with more fine details. For the motion prediction task, we utilize
the semantic segmentations produced by predictive parsing to separately estimate
motion for pixels with different categories. In order to perform the results for multiple
time steps, we take the predictions as input and iterate the model to predict
subsequent frames. The proposed model has a generic framework which is agnostic to
backbone deep networks and can be conveniently trained in an end-to-end manner.

Taking Cityscapes~\cite{cityscape} as testbed, we conduct extensive experiments
to verify the effectiveness of our model in future prediction. Our model
significantly improves mIoU of parsing predictions and reduces the endpoint
error (EPE) of flow predictions compared to strongly competitive baselines
including a warping method based on optical flow, standalone parsing prediction or
flow prediction and other state-of-the-arts methods~\cite{s2s}. We also present
how to predict steering angles using the proposed model.

\section{Related work}
For the general field of classic flow (motion) estimation and image semantic
segmentation, which is out of this paper's scope, we refer the readers to
comprehensive review articles~\cite{baker2011database,garcia2017review}. Below
we mainly review existing works that focus on predictive tasks.

\paragraph{Flow and scene parsing prediction}
The research on predictive scene parsing or motion prediction is still relatively under-explored. All existing works in this direction tackle the parsing
prediction and flow prediction as independent tasks. With regards to motion prediction, Luo \textit{et
  al.}~\cite{luo2017} employed a convolutional LSTM architecture to predict
sequences of 3D optical flow. Walker \textit{et al.}~\cite{walker2014patch} made
long-term motion and appearance prediction via a transition and context
model. \cite{srivastava2015unsupervised} trained CNN for predicting motion of
handwritten characters in a synthetic dataset. \cite{denseof} predicted
future optical flow given a static image. Different from above works, our model
not only predicts the flow but also scene parsing at the same time,
which definitely provides richer information to visual systems.

There are also only a handful number of works exploring the prediction of scene parsing in future frames. Jin
\textit{et al.}~\cite{jinme} trained a deep model to predict the segmentations
of the next frame from preceding input frames, which is shown to be beneficial
for still-image parsing task. Based on the network proposed in \cite{mathieu},
Natalia \textit{et al.}~\cite{s2s} predicted longer-term parsing maps for future
frames using the preceding frames' parsing maps. Different from~\cite{s2s}, we
simultaneously predict optical flows for future frames. Benefited from the
discriminative local features learned from flow prediction, the model produces
more accurate parsing results. Another related work to ours
is~\cite{patraucean2015spatio} which employed an RNN to predict the optical flow
and used the flow to warp preceding segmentations. Rather than simply producing
the future parsing map through warping, our model predicts flow and scene
parsing jointly using learning methods. More importantly, we leverage the
benefit that each task brings to the other to produce better results for both
flow prediction and parsing prediction.

\paragraph{Predictive learning}
While there are few works specifically on predictive scene parsing or dense motion prediction, learning to prediction in general has received a significant attention from the research community in recent years. Research
in this area has explored different aspects of this problem. \cite{yuen2010data}
focused on predicting the trajectory of objects given input
image. \cite{hoai2014max} predicted the action class in the future
frames. Generative adversarial networks (GAN) are firstly introduced
in~\cite{goodfellowgan} to generate natural images from random noise, and have
been widely used in many fields including image synthesis \cite{goodfellowgan},
future prediction
\cite{lotter2015unsupervised,mathieu,vondrick2016generating,denseof,villegas2017decomposing,villegas2017learning} and
semantic inpainting~\cite{trevor16inpainting}. Different from above methods, our
model explores a new predictive task, \textit{i.e.} predicting the scene parsing
and motion dynamics in the future simultaneously.

\paragraph{Multi-task learning}
Multi-task learning~\cite{argyriou2007multi,evgeniou2004regularized} aims to solve multiple tasks jointly by taking advantage of the shared domain knowledge in related tasks. Our work is partially related to multi-task learning in that both the parsing results and motion dynamics are predicted jointly in a single model. However, we note that predicting parsing and motion ``in the future'' is a novel and challenging task which cannot be straightforwardly tackled by conventional multi-task learning methods. To our best knowledge, our work is the first solution to this challenging task.

\begin{figure}[t!]
\captionsetup[subfigure]{labelformat=empty}
	\centering
	\subfloat[]{
	    \includegraphics[width=0.8\linewidth]{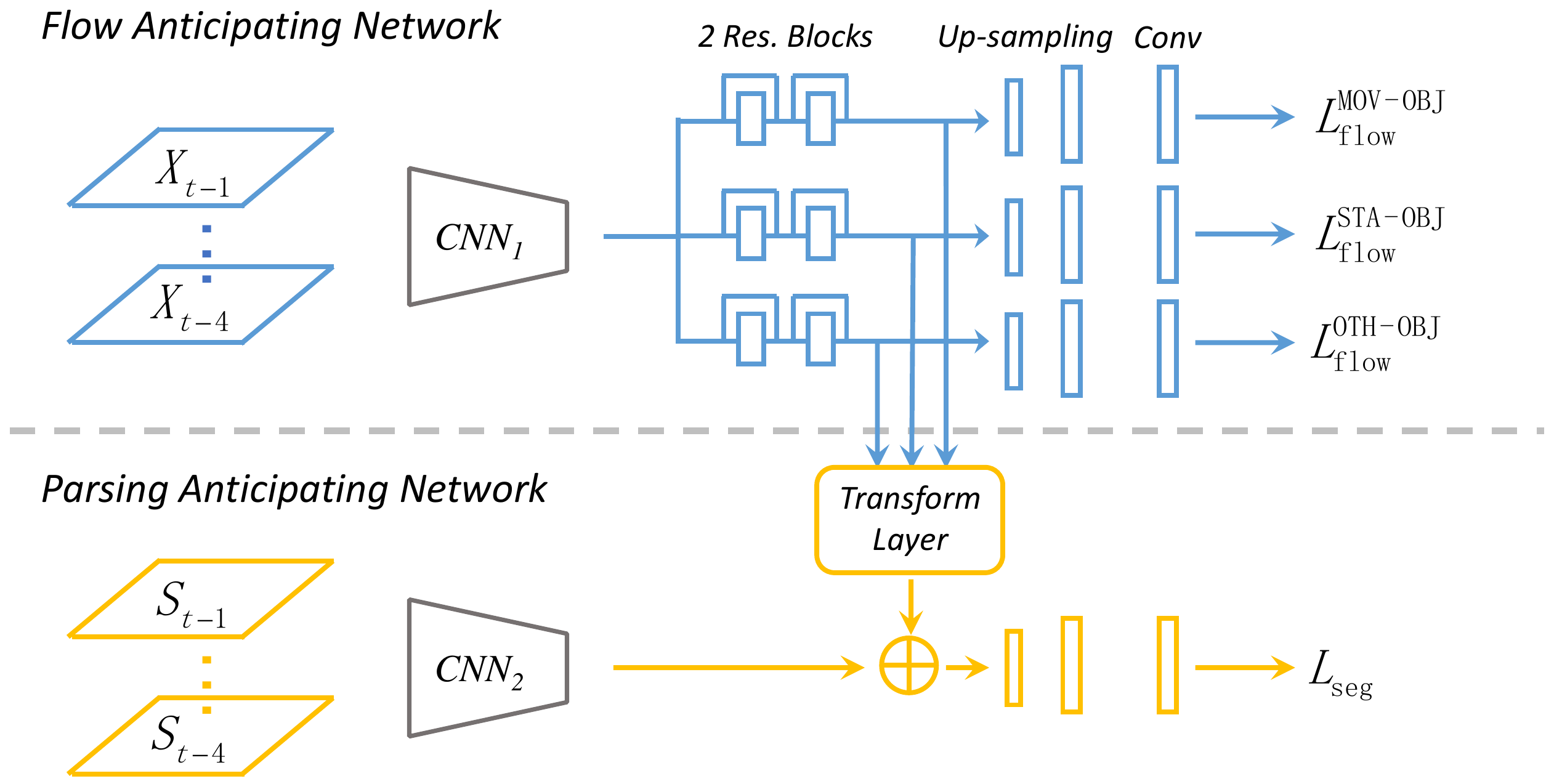}%
	  }
          \caption{The framework of our model for predicting future scene
            parsing and optical flow for one time step ahead. Our model is motivated by the assumption that flow and parsing prediction are mutually beneficial. We design the architecture to promote such mutual benefits. The model consists
            of two module networks, \textit{i.e.} the flow anticipating network
            (blue) which takes preceding frames: $X_{t-4:t-1}$ as input and predicts future flow and the
            parsing anticipating network (yellow) which takes the preceding parsing results: $S_{t-4:t-1}$ as input and predicts future scene
            parsing. By providing pixel-level class information (\textit{i.e.} $S_{t-1}$), the parsing anticipating network benefits the flow anticipating network to enable the latter to semantically distinguish different pixels (\textit{i.e.} moving/static/other objects) and predict their flows more accurately in the corresponding branch. Through the transform layer, the discriminative local
            features learned by the flow anticipating network are combined with
            the parsing anticipating network to facilitate parsing over small objects and avoid over-smooth in parsing predictions. When predicting multiple time-steps ahead, the prediction of the parsing network in a time-step is used as the input in the next time-step.}
    \label{fig:framework}
\end{figure}

\section{Predicting scene parsing and motion dynamics in the future}
\label{sec:train}
In this section, we first propose our model for predicting semantics and motion dynamics one time step ahead, and then extend our model to perform predictions for multiple time steps.

Due to high cost of acquiring dense human annotations of optical flow and
scene parsing for natural scene videos, only subset of frames are labeled for scene parsing in the current datasets. Following~\cite{s2s}, to circumvent the need for datasets with dense
annotations, we train an adapted Res101 model (denoted as Res101-FCN, more
details are given in Sec.~\ref{sec:setting}) for scene parsing to produce the target
semantic segmentations for frames without human annotations. Similarly, to
obtain the dense flow map for each frame, we use the output of the
state-of-the-art epicflow~\cite{epicflow} as our target optical flow. Note that our model is orthogonal to specific flow methods since they are only used to produce the target flow for training the flow anticipating network. Notations used
in the following text are as follows. $X_i$ denotes the $i$-th frame of a video
and $X_{t-k:t-1}$ denotes the sequence of frames with length $k$ from $X_{t-k}$
to $X_{t-1}$. The semantic segmentation of $X_t$ is denoted as $S_t$, which is
the output of the penultimate layer of Res101-FCN. $S_t$ has the same spatial
size as $X_t$ and is a vector of length $C$ at each location, where $C$ is the
number of semantic classes. We denote $O_{t}$ as the pixel-wise optical flow map
from $X_{t-1}$ to $X_t$, which is estimated via
epicflow~\cite{epicflow}. Correspondingly, $\hat S_t$ and $\hat O_t$ denote the
predicted semantic segmentation and optical flow.

\subsection{Prediction for one time step ahead}
\paragraph{Model overview}
The key idea of our approach is to model flow prediction and parsing prediction
jointly, which are potentially mutually beneficial. As illustrated in
Figure~\ref{fig:framework}, the proposed model consists of two module networks
that are trained jointly, \textit{i.e.} the flow anticipating network that
takes preceding frames $X_{t-k:t-1}$ as input to output the pixelwise flow
prediction for $O_{t}$ (from $X_{t-1}$ to $X_t$), and the parsing anticipating
network that takes the segmentation of preceding frames $S_{t-k:t-1}$ as input
to output pixelwise semantic prediction for an unobserved frame $X_{t}$. The
mutual influences of each network on the other are exploited in two
aspects. First, the last segmentations $S_{t-1}$ produced by the parsing
anticipating network convey pixel-wise class labels, which are used by the flow
anticipating network to predict optical flow values for each pixel according to
its belonging object group, \textit{e.g.} moving objects or static
objects. Second, the parsing anticipating network combines the discriminative
local feature learned by the flow anticipating network to produce sharper and
more accurate parsing predictions.

Since both parsing prediction and flow prediction are essentially both the
dense classification problem, we use the same deep architecture (Res101-FCN) for
predicting parsing results and optical flow. Note the Res101-FCN used in this
paper can be replaced by any CNNs. We adjust the input/output layers of these
two networks according to the different channels of their input/output. The
features extracted by feature encoders (CNN$_1$ and CNN$_2$) are spatially
enlarged via up-sampling layers and finally fed to a convolutional layer to
produce pixel-wise predictions which have the same spatial size as input.

\paragraph{Flow anticipating network}
In videos captured for autonomous driving or navigation, regions with
different class labels have different motion patterns. For example, the motion
of static objects like \textit{road} is only caused by the motion of the camera
while the motion of moving objects is a combination of motions from both the camera
and objects themselves. Therefore compared to methods that predict all pixels'
optical flow in a single output layer, it would largely reduce the difficulty of
feature learning by separately modeling the motion of regions with different
classes. Following~\cite{3flow}, we assign each class into one of three
pre-defined object groups, \textit{i.e.}
$\mathcal{G=\{\textit{moving objects (MOV-OBJ)}, \textit{static objects
    (STA-OBJ)}, \textit{other objects (OTH-OBJ)}\}}$
in which MOV-OBJ includes pedestrians, truck, \textit{etc.}, STA-OBJ includes
sky, road, \textit{etc.}, and OTH-OBJ includes vegetation and buildings,
\textit{etc.} which have diverse motion patterns and shapes. We append a small network
(consisting of two residual blocks) to the feature encoder (CNN$_1$) for each
object group to learn specified motion representations. During training, the
loss for each pixel is only generated at the branch that corresponds to the
object group to which the pixel belongs. Similarly, in testing, the flow
prediction for each pixel is generated by the corresponding branch. The loss
function between the model output $\hat O_t$ and target output $O_t$ is

\begin{equation}
  L_{\text{flow}}(\hat O_t,O_t)  = \sum\limits_{g\in \mathcal{G} } L_{\text{flow}}^{g}; \ \ \ \ L_{\text{flow}}^{g} = {\frac{1}{{\left| {N_g } \right|}}} \sum\limits_{(i,j) \in N_g } {\left\| {O_t^{i,j}  - \hat O_t^{i,j}} \right\|_2 }
\end{equation}

\noindent where $(i,j)$ index the pixel in the region $N_g$.

\paragraph{Parsing anticipating network}
\label{sec:pan}
The input of the parsing anticipating network is a sequence of preceding
segmentations $S_{t-k:t-1}$. We also explore other input space alternatives,
including preceding frames $X_{t-k:t-1}$, and the combination of preceding
frames and corresponding segmentations $X_{t-k:t-1}S_{t-k:t-1}$, and we observe
that the input $S_{t-k:t-1}$ achieves the best prediction performance. We
conjecture it is easier to learn the mapping between variables in the same
domain (\textit{i.e.} both are semantic segmentations). However, there are two
drawbacks brought by this strategy. Firstly, $S_{t-k:t-1}$ lose the
discriminative local features \textit{e.g.} color, texture and shape
\textit{etc.}, leading to the missing of small objects in predictions, as
illustrated in Figure~\ref{fig:single} (see yellow boxes). The flow prediction
network may learn such features from the input frames. Secondly, due to the lack
of local features in $S_{t-k:t-1}$, it is difficult to learn accurate pixel-wise
correspondence in the parsing anticipating network, which causes the predicted
labeling maps to be over-smooth, as shown in Figure~\ref{fig:single}. The flow
prediction network can provide reliable dense pixel-wise correspondence by
regressing to the target optical flow. Therefore, we integrate
the features learned by the flow anticipating network with the parsing prediction
network through a transform layer (a shallow CNN) to improve the quality of
predicted labeling maps. Depending on whether human annotations are available,
the loss function is defined as

\begin{equation}
 \footnotesize
  \begin{aligned}
    \label{eq:loss_seg}
    L_{\text{seg}} (\hat S, S) =
    \begin{cases}
      - \sum\limits_{(i,j) \in X_t } {{\log (\hat S_t^{i,j} (c))} } , & X_t \ \ \text{has human annotation}, \\
      L_{\ell _1 }(\hat S, S) + L_{\text{gdl}}(\hat S, S), & \text{otherwise}
    \end{cases}
  \end{aligned}
\end{equation}

\noindent where $c$ is the ground truth class for the pixel at location
$(i,j)$. It is a conventional pixel-wise cross-entropy loss when $X_t$ has human
annotations. $L_{\ell _1 }$ and $L_{\text{gdl}}$ are $\ell_1$ loss and gradient
difference loss~\cite{mathieu} which are defined as
\begin{equation*} 
\footnotesize
\begin{split}
  L_{\ell _1 }(\hat S, S) &= \sum\limits_{(i,j) \in X_t } {\left| {S_t^{i,j}  - \hat S_t^{i,j} } \right|}, \\
  L_{\text{gdl}} &= \sum\limits_{(i,j) \in X_t } {\left(\left| {|{S_t^{i,j} -
            S_t^{i - 1,j} }| - |{\hat S_t^{i,j} - \hat S_t^{i - 1,j} } |}
      \right| + \left| {|{S_t^{i,j - 1} - S_t^{i,j} }| - |{\hat S_t^{i,j - 1} -
            \hat S_t^{i,j} }|} \right|\right)}.
\end{split}
\end{equation*}
The $\ell_1$ loss encourages predictions to regress to the target values while
the gradient difference loss produces large errors in the gradients of the
target and predictions. 

The reason for using different losses for human and non-human annotated frames in Eq.~\ref{eq:loss_seg} is that the automatically produced parsing ground-truth (by the pre-trained Res101-FCN) of the latter may contain wrong annotations. The cross-entropy loss using one-hot vectors as labels is sensitive to the wrong annotations. Comparatively, the ground-truth labels used in the combined loss ($L_{\ell _1 }+ L_{\text{gdl}}$) are inputs of the softmax layer (ref. Sec.~\ref{sec:train}) which allow for non-zero values in more than one category, thus our model can learn useful information from the correct category even if the annotation is wrong. We find replacing $L_{\ell _1 }+ L_{\text{gdl}}$ with the cross-entropy loss reduces the mIoU of the baseline S2S (\textit{i.e.} the parsing participating network) by 1.5 from 66.1 when predicting the results one time-step ahead.

Now we proceed to explain the role of the transform
layer which transforms the features of CNN$_1$ before combining them with those
of CNN$_2$. Compared with naively combining the features from two networks
(\textit{e.g.}, concatenation), the transform layer brings the following two
advantages: 1) naturally normalize the feature maps to proper scales; 2) align
the features of semantic meaning such that the integrated features are more
powerful for parsing prediction. Effectiveness of this transform layer is
clearly validated in the ablation study in Sec.~\ref{sec:analysis}.

The final objective of our model is to minimize the combination of losses from the
flow anticipating network and the parsing anticipating network as follows

\begin{equation*}
  L(X_{t-k:t-1},S_{t-k:t-1},\hat X_t, \hat S_t) = L_{\text{flow}}(\hat O_t,O_t) + L_{\text{seg}}(\hat S, S).
\end{equation*}

\subsection{Prediction for multiple time steps ahead}
\label{sec:pmt}
Based on the above model which predicts scene parsing and flow for the single future time step, we explore two
ways to predict further into the future. Firstly, we iteratively apply the model
to predict one more time step into the future by treating the prediction as
input in a recursive way. Specifically, for predicting multiple time steps in the flow anticipating
network, we warp the most recent frame $X_{t-1}$ using the output prediction
$\hat O_{t}$ to get the $\hat X_t$ which is then combined with $X_{t-k-1:t-1}$
to feed the flow anticipating network to generate $\hat O_{t+1}$, and so
forth. For the parsing anticipating network, we combine the predicted parsing
map $\hat S_t$ with $S_{t-k-1:t-1}$ as the input to generate the parsing
prediction at $t+1$. This scheme is easy to implement and allows us to
predict arbitrarily far into the future without increasing training complexity w.r.t.
with the number of time-steps we want to predict. Secondly, we fine-tune our
model by taking into account the influence that the recurrence has on prediction
for multiple time steps. We apply our model recurrently as described above to
predict two time steps ahead and apply the back propagation through time
(BPTT)~\cite{jaeger2002tutorial} to update the weight. We have verified through
experiments that the fine-tuning approach can further improve the performance as
it models longer temporal dynamics during training.  
\section{Experiment}
\subsection{Experimental settings}
\label{sec:setting}
\begin{figure}[t!]
\captionsetup[subfigure]{labelformat=empty}
	\centering
	\subfloat[]{%
	    \includegraphics[width=\linewidth]{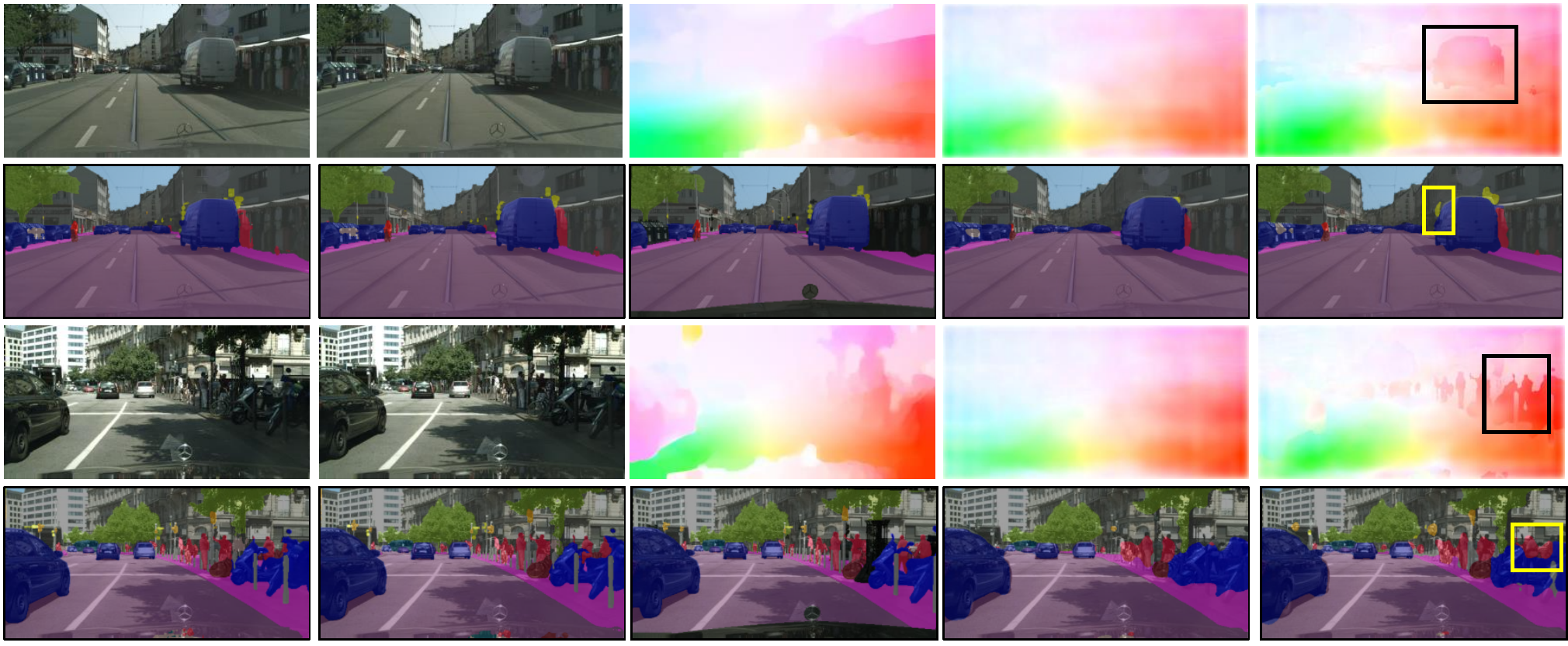}%
	  }
          \caption{Two examples of prediction results for predicting one time
            step ahead. \textbf{Odd row}: The images from left to right are
            $X_{t-2}$, $X_{t-1}$, the target optical flow map $O_{t}$, the flow
            predictions from PredFlow and the flow predictions from our
            model. \textbf{Even row}: The images from left to right are
            $S_{t-2}$, $S_{t-1}$, the ground truth semantic annotations at the
            time $t$, the parsing prediction from $S2S$ and the parsing
            prediction from our model. The flow predictions from our model show
            clearer object boundaries and predict more accurate values for
            moving objects (see black boxes) compared to PredFlow. Our model is
            superior to S2S by being more discriminative to the small objects in
            parsing predictions (see yellow boxes). }
    \label{fig:single}
\end{figure}

\begin{figure}[t!]
\captionsetup[subfigure]{labelformat=empty}
	\centering
	\subfloat[]{%
	    \includegraphics[width=\linewidth]{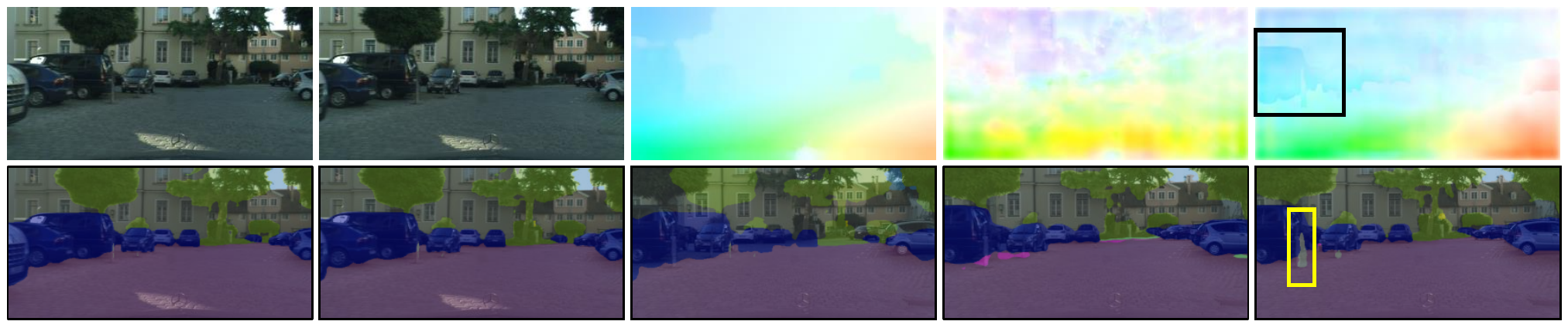}%
	  }
          \caption{An example of prediction results for predicting ten time
            steps ahead. \textbf{Top} (from left to right): $X_{t-11}$,
            $X_{t-10}$, the target optical flow map $O_t$, the flow prediction
            from PredFlow and the flow prediction from our model.
            \textbf{Bottom} (from left to right): $S_{t-11}$, $S_{t-10}$, the
            ground truth semantic annotation at the time $t$, the parsing
            prediction from $S2S$ and the parsing prediction from our model. Our
            model outputs better prediction compared to PredFlow (see black
            boxes) and S2S (see yellow boxes).}
    \label{fig:multi}
\end{figure}
\paragraph{Datasets}
We verify our model on the large scale Cityscapes~\cite{cityscape} dataset
which contains $2{,}975/500$ train/val video sequences with $19$
semantic classes. Each video sequence lasts for 1.8$s$ and contains $30$ frames,
among which the $20$th frame has fine human annotations. Every frame in
Cityscapes has a resolution of $1{,}024\times2{,}048$ pixels.
\paragraph{Evaluation criteria}
We use the mean IoU (mIoU) for evaluating the performance of predicted parsing
results on those $500$ frames in the val set with human annotations. For
evaluating the performance of flow prediction, we use the average endpoint error
(EPE)~\cite{baker2011database} following conventions~\cite{flownet1} which is
defined as $\frac{1}{N}\sqrt {(u - u_{\text{GT}} )^2 + (v - v_{\text{GT}} )^2 }$
where $N$ is the number of pixels per-frame, and $u$ and $v$ are the components
of optical flow along $x$ and $y$ directions, respectively. To be consistent
with mIoU, EPEs are also reported on the $20$th frame in each val sequence.
\paragraph{Baselines}
To fully demonstrate the advantages of our model on producing better
predictions, we compare our model against the following baseline methods:

\begin{itemize}
  \item \textit{Copy last input}\quad Copy the last optical flow ($O_{t-1}$) and
    parsing map ($S_{t-1}$) at time $t-1$ as predictions at time $t$.
  \item \textit{Warp last input}\quad Warp the last segmentation $S_{t-1}$ using
    $O_{t-1}$ to get the parsing prediction at the next time step. In order to
    make flow applicable to the correct locations, we also warp the flow field
    using the optical flow in each time step.
  \item \textit{PredFlow}\quad Perform flow prediction without the object masks
    generated from segmentations. The architecture is the same as the flow
    prediction net in Figure~\ref{fig:framework} which generates pixel-wise flow
    prediction in a single layer, instead of multiple branches. For fair
    comparison with our joint model, in the following we report the average
    result of two independent \textit{PredFlow} with different random
    initializations. When predicting the segmentations at time $t$, we use the
    flow prediction output by \textit{PredFlow} at time $t$ to warp the
    segmentations at time $t-1$. This baseline aims to verify the advantages
    brought by parsing prediction when predicting flow.
  \item \textit{S2S}~\cite{s2s}\quad Use only parsing anticipating
    network. The difference is that the former does not leverage features
    learned by the flow anticipating network to produce parsing predictions. We
    replace the backbone network in the original S2S as the same one of ours,
    \textit{i.e.} Res101-FCN and retrain S2S with the same configurations as
    those of ours. Similar to the \textit{PredFlow}, the average performance of
    two randomly initialized \textit{S2S} is reported. This baseline aims to
    verify the advantages brought by flow prediction when predicting parsing.
\end{itemize}
\paragraph{Implementation details}
Throughout the experiments, we set the length of the input sequence as 4
frames, \textit{i.e.} $k=4$ in $X_{t-k:t-1}$ and $S_{t-k:t-1}$
(ref. Sec.~\ref{sec:train}). The original frames are firstly downsampled to the resolution of
$256\times512$ to accelerate training. In the flow anticipating network, we assign
19 semantic classes into three object groups which are defined as follows:
MOV-OBJ including person, rider, car, truck, bus, train, motorcycle and bicycle, STA-OBJ including road, sidewalk, sky, pole, traffic light and traffic sign
and OTH-OBJ including building, wall, fence, terrain and vegetation. For data
augmentation, we randomly crop a patch with the size of $256\times 256$ and
perform random mirror for all networks. All results of our model are based on
single-model single-scale testing. For other hyperparameters including weight
decay, learning rate, batch size and epoch number \textit{etc}., please refer to
the supplementary material. All of our experiments are carried out on NVIDIA Titan
X GPUs using the Caffe library. 

\subsection{Results and analysis} Examples of the flow predictions and parsing
predictions output by our model for one-time step and ten-time steps are
illustrated in Figure~\ref{fig:single} and Figure~\ref{fig:multi}
respectively. Compared to baseline models, our model produces more visually
convincing prediction results.
\begin{table}
\begin{minipage}[l]{0.49\textwidth}%
  \caption{The performance of parsing prediction on Cityscapes val set. For each
    competing model, we list the mIoU/EPE when predicting one time step
    ahead. Best results in bold.}
  \label{table:onetime}
  \centering
  \begin{tabular}{lcc} \toprule Model  & {mIoU}       &{EPE}  \\
    \midrule
    Copy last input             & 59.7            & 3.03   \\
    Warp last input             & 61.3            & 3.03   \\
    PredFlow                    & 61.3            & 2.71   \\
    S2S~\cite{s2s}                      & 62.6            & -   \\
    \midrule
    ours (w/o Trans. layer)      & 64.7            & 2.42 \\
    ours                        & \bf{66.1}            & \bf{2.30}   \\
    \bottomrule
  \end{tabular}
\end{minipage}\ \ \
\begin{minipage}[r]{0.49\textwidth}
  \caption{The performance of motion prediction on Cityscapes val set. For each
    model, we list the mIoU/EPE when predicting one time step ahead. Best
    results in bold.}
  \label{table:multime}
  \centering
  \begin{tabular}{lcc} \toprule Model  & {mIoU}       &{EPE}  \\
    \midrule
    Copy last input             & 41.3            & 9.40   \\
    Warp last input             & 42.0            & 9.40   \\
    PredFlow                    & 43.6            & 8.10   \\
    S2S~\cite{s2s}                      & 50.8            &   - \\
    \midrule
    ours (w/o Recur. FT)        & 52.6            & 6.63   \\
    ours                        & \bf{53.9}            & \bf{6.31}   \\
    \bottomrule
    \end{tabular}
\end{minipage}
\end{table}

\subsubsection{One-time step anticipation}
\label{sec:analysis}
Table~\ref{table:onetime} lists the performance of parsing and flow
prediction on the $20$th frame in the val set which has ground truth semantic
annotations. It can be observed that our model achieves the best performance on
both tasks, demonstrating the effectiveness on learning
the latent representations for future prediction. Based on the results, we analyze the
effect of each component in our model as follows.

\textbf{The effect of flow prediction on parsing prediction}\quad Compared with
S2S which does not leverage flow predictions, our model improves the mIoU with a large margin (3.5\%). As shown in
Figure~\ref{fig:single}, compared to S2S, our model performs better on localizing the small objects in the predictions \textit{e.g.} pedestrian and
traffic sign, because it combines the discriminative local features learned in
the flow anticipating network. These results clearly demonstrate the benefit of flow
prediction for parsing prediction.

\textbf{The effect of parsing prediction on flow prediction}\quad Compared with
the baseline PredFlow which has no access to the semantic information when
predicting the flow, our model reduces the average EPE from 2.71 to 2.30 (a 15\%
improvement), which demonstrates parsing prediction is beneficial to flow
prediction. As illustrated in Figure~\ref{fig:single}, the improvement our model
makes upon PredFlow comes from two aspects. First, since the segmentations
provide boundary information of objects, the flow map predicted by our model has
clearer object boundaries while the flow map predicted by PredFlow is mostly
blurry. Second, our model shows more accurate flow predictions on the moving
objects (ref. Sec.~\ref{sec:setting} for the list of moving objects). We
calculate the average EPE for only the moving objects, which is 2.45 for our
model and 3.06 for PredFlow. By modeling the motion of different objects
separately, our model learns better representation for each motion mode. If all
motions are predicted in one layer as in PredFlow, then the moving objects which
have large displacement than other regions are prone to smoothness.

\textbf{Benefits of the transform layer} As introduced in Sec.~\ref{sec:pan},
the transform layer improves the performance of our model by learning the latent
feature space transformations from CNN$_1$ to CNN$_2$. In our experiments, the
transform layer contains one residual block~\cite{residual} which has been
widely used due to its good performance and easy optimization. Details of the
residual block used in our experiments are included in the supplementary
material. Compared to the variant of our model w/o the transform layer, adding
the transform layer improves the mIoU by 1.4 and reduces EPE by 0.12. We observe
that stacking more residual blocks only leads to marginal improvements at larger
computational costs.  
\subsubsection{Longer duration prediction}
The comparison of the prediction performance among all methods for ten time
steps ahead is listed in Table~\ref{table:multime}, from which one can observe
that our model performs the best in this challenging task. The effect of each
component in our model is also verified in this experiment. Specifically,
compared with S2S, our model improves the mIoU by 3.1\% due to the synergy with
the flow anticipating network. The parsing prediction helps reducing the EPE of PredFlow by 1.79. Qualitative results are illustrated in
Figure~\ref{fig:multi}.

\textbf{The effect of recurrent fine-tuning} As explained in Sec.~\ref{sec:pmt},
it helps our model to capture long term video dynamics by fine-tuning the
weights when recurrently applying the model to predict the next time step in the
future. As shown in Table~\ref{table:multime}, compared to the variant w/o
recurrent ft, our model w/ recurrent fine-tuning improves the mIoU by 1.3\% and
reduces the EPE by 0.32, therefore verifying the effect of recurrent
fine-tuning. 
\subsection{Application for predicting the steering angle of a vehicle}
\begin{wraptable}{r}{6.5cm}
	\caption{Comparison results of steering angle prediction on a dataset from Comma.ai~\cite{commaai}. The criteria is the mean square error (MSE, in \textit{degree}$^2$) between the prediction and groud truth.}\label{steering}
	\small
	\begin{tabular}{L{3.1cm}cc}\\\toprule
		Model & MSE (\textit{degrees}$^2$) \\\midrule
		Copy last prediction & 4.81\\
		Comma.ai\footnotemark~\cite{commaai} & $\sim$ 4\\  \midrule
		ours & 2.96 \\  \bottomrule
		\label{tabel:steer}
	\end{tabular}
\end{wraptable}
\footnotetext{\scriptsize{$\textrm{https://}\textrm{github}.\textrm{com/}\textrm{commaai/}\textrm{research}$}}
With the parsing prediction and flow prediction available, one can enable the
moving agent to be more alert about the environments and get ``smarter''. Here, we
investigate one application: predicting the steering angle of the
vehicle. The intuition is it is convenient to infer the steering angle given the
predicted flow of static objects, \textit{e.g.} road and sky, the motion of
which is only caused by ego-motion of the camera mounted on the
vehicle. Specifically, we append a fully connected layer to take the features
learned in the STA-OBJ branch in the flow anticipating network as input and
perform regression to steering angles. We test our model on the dataset from
Comma.ai~\cite{commaai} which consists of 11 videos amounting to about 7
hours. 
The data of steering angles have been recorded for each frame captured at 20Hz with the resolution of $160\times 320$. 
We randomly sample
50K/5K frames from the train set for training and validation purpose. Since
there are videos captured at night, we normalize all training frames to
$[0, 255]$. Similar to Cityscapes, we use epicflow and Res101-FCN to produce the
target output for flow prediction and parsing prediction, respectively. 
We first
train our model following Sec.~\ref{sec:train} and then fine-tune the whole
model with the MSE loss after adding the fully connected layer for steering
angle prediction. 
During training, random crop with the size of $160\times 160$
and random mirror are employed and other hyperparameter settings follow
Sec.~\ref{sec:setting}. The testing results are listed in
Table~\ref{tabel:steer}. Compared to the model from Comma.ai which uses a five-layer CNN to
estimate the steering angle from a single frame and is trained end-to-end on all the training frames (396K), our model achieves much better performance (2.84 versus $\sim$4
in \textit{degrees}$^2$). Although we do not push the performance by using more
training data and more complex prediction models (only a fully connected layer
is used in our model for output steering angle), this preliminary experiment
still verifies the advantage of our model in learning the underlying latent
parameters. We think it is just an initial attempt in validating the dense prediction results through applications, which hopefully can stimulate researchers to explore other interesting ways to
utilize the parsing prediction and flow prediction.  

\section{Conclusion}
In this paper, we proposed a novel model to predict the future scene parsing and
motion dynamics. To our best knowledge, this is the first research attempt to
anticipate visual dynamics for building intelligent agents. The model consists
of two networks: the flow anticipating network and the parsing
anticipating network which are jointly trained and benefit each other. On the
large scale Cityscapes dataset, the experimental results demonstrate that the
proposed model generates more accurate prediction than well-established baselines
both on single time step prediction and multiple time prediction. In addition,
we also presented a method to predict the steering angle of a vehicle using our model
and achieve promising preliminary results on the task.  

\paragraph{Acknowledgements} The work of Jiashi Feng was partially supported by National University of Singapore startup grant
R-263-000-C08-133, Ministry of Education of Singapore AcRF Tier One grant R-263-000-C21-112
and NUS IDS grant R-263-000-C67-646.

\bibliographystyle{plain}
\footnotesize \bibliography{mybibfile}



\section*{Supplementary material}

# Predicting Scene Parsing and Motion Dynamics in the Future
# (Supplementary Material)

**Xiaojie Jin[1], Huaxin Xiao[2], Xiaohui Shen[3], Jimei Yang[3], Zhe Lin[3]**
**Yunpeng Chen[2], Zequn Jie[4], Jiashi Feng[2], Shuicheng Yan[5,2]**
[1]NUS Graduate School for Integrative Science and Engineering (NGS), NUS
[2]Department of ECE, NUS   [3]Adobe Research   [4]Tencent AI Lab   [5]Qihoo 360 AI Institute

## Abstract

In supplementary material, we provide the architecture details and training settings of Res101-FCN and the transform layer.

## 1 Implementation Details

### 1.1 Network Architecture

The architecture of our Res101-FCN is illustrated in Figure 1. It is modified from the original Res101 [2] by adapting it to a fully convolutional network, following [3]. Specifically, we replace the average pooling layer and the 1,000-way classification layer with a fully convolutional layer (denoted as conv5_3$_{\text{cls}}$ in Figure 1) to produce dense parsing maps. Also, we modify conv5_1, conv5_2 and conv5_3 to dilated convolutional layers by setting their dilation size to 2 to enlarge the size of the receptive field. Then, the output feature map of conv5_3 has a stride of 16. Following FCN [3], we utilize high-frequency features learned in bottom layers by adding skip connections from conv1, pool1, conv3_3 to corresponding up-sampling to produce parsing maps with the same size as input frames. The training setting are listed at Table 1. The mIoU of Res101-FCN on val of Cityscapes is 75.2%.

### 1.2 Architecture of Transform Layer

In our experiments, a "bottleneck" residual block proposed in [2] is used as the transform layer whose architecture is illustrated in Figure 2.

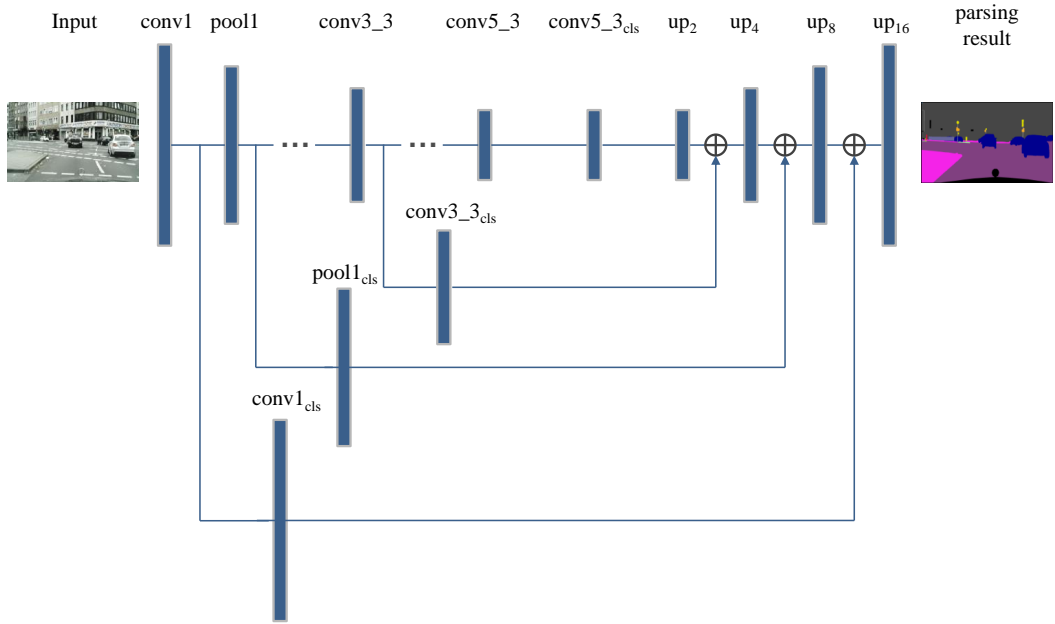

Figure 1: Architecture of Res101-FCN used in our experiments. The subscript "cls" means the corresponding layer is a convolutional layer which has the same number of filters as the categorical classes, *e.g.* the `conv5_3`$_{\text{cls}}$ that has 19 filters for Cityscapes dataset. Following FCN [3], skip connections from `conv1`, `pool1`, `conv3_3` are constructed to utilize high frequency information in bottom layers. The `up`$_n$ is an up-sampling layer, of which the output feature map is $n$ times the spatial size of that of `conv5_3`$_{\text{cls}}$. The symbol $\oplus$ represents the operation of "summation".

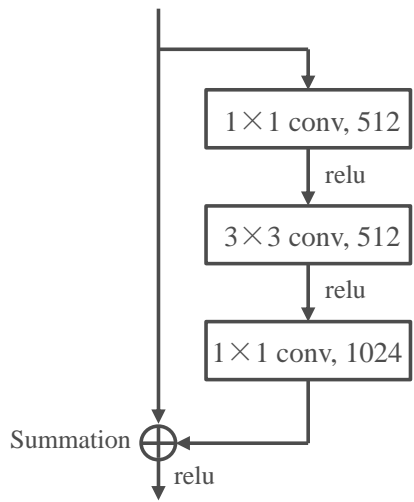

Figure 2: Architecture of the residual block [2] used as the transform layer in our model. In the residual block, the shortcut connection performs identity mapping, the output of which is added to the output of the stacked convolutional layers.

Table 1: Training settings of Res101-FCN. "LR" stands for "learning rate". In all experiments, the weight decay and momentum are set to 0.0001 and 0.9 respectively. Following Deeplab [1], the "poly" learning rate policy is used and the value of power is equal to 0.9.

| Batch size | Crop size | Overall epochs | LR-initial | LR-policy |
|---|---|---|---|---|
| 4 | 256 | 30 | 1e-4 | Poly |

2


\end{document}